\documentclass[letterpaper]{article} 
\usepackage{aaai2026}  
\usepackage{times}  
\usepackage{helvet}  
\usepackage{courier}  
\usepackage[hyphens]{url}  
\usepackage{graphicx} 
\usepackage{natbib}  
\usepackage{caption} 
\usepackage{algorithm}
\usepackage{algorithmic}

\usepackage{newfloat}
\usepackage{listings}
\DeclareCaptionStyle{ruled}{labelfont=normalfont,labelsep=colon,strut=off} 
\floatstyle{ruled}
\newfloat{listing}{tb}{lst}{}
\floatname{listing}{Listing}
\usepackage{array}
\usepackage{multirow}
\usepackage{xcolor}
\usepackage{colortbl}
\usepackage{bm}
\usepackage{pifont}
\usepackage{subfig}
\usepackage[accsupp]{axessibility}
\usepackage{hhline}

\usepackage{url}

\title{CKDA: Cross-modality Knowledge Disentanglement and Alignment\\for Visible-Infrared Lifelong Person Re-identification}
\author{
    Zhenyu Cui,
    Jiahuan Zhou,
    Yuxin Peng\thanks{Corresponding author.}
}
\affiliations{
    Wangxuan Institute of Computer Technology, Peking University\\

    cuizhenyu@stu.pku.edu.cn, \{jiahuanzhou, pengyuxin\}@pku.edu.cn
}

\usepackage{bibentry}

\begin{document}

\maketitle

\begin{abstract}
Lifelong person Re-IDentification (LReID) aims to match the same person employing continuously collected individual data from different scenarios. To achieve continuous all-day person matching across day and night, Visible-Infrared Lifelong person Re-IDentification (VI-LReID) focuses on sequential training on data from visible and infrared modalities and pursues average performance over all data. To this end, existing methods typically exploit cross-modal knowledge distillation to alleviate the catastrophic forgetting of old knowledge. However, these methods ignore the mutual interference of modality-specific knowledge acquisition and modality-common knowledge anti-forgetting, where conflicting knowledge leads to collaborative forgetting. To address the above problems, this paper proposes a Cross-modality Knowledge Disentanglement and Alignment method, called CKDA, which explicitly separates and preserves modality-specific knowledge and modality-common knowledge in a balanced way. Specifically, a Modality-Common Prompting (MCP) module and a Modality-Specific Prompting (MSP) module are proposed to explicitly disentangle and purify discriminative information that coexists and is specific to different modalities, avoiding the mutual interference between both knowledge. In addition, a Cross-modal Knowledge Alignment (CKA) module is designed to further align the disentangled new knowledge with the old one in two mutually independent inter- and intra-modality feature spaces based on dual-modality prototypes in a balanced manner. Extensive experiments on four benchmark datasets verify the effectiveness and superiority of our CKDA against state-of-the-art methods. The source code of this paper is available at \url{https://github.com/PKU-ICST-MIPL/CKDA-AAAI2026}.
\end{abstract}


\section{Introduction}

\begin{figure}[ht]
\centering
\includegraphics[width=0.96\linewidth]{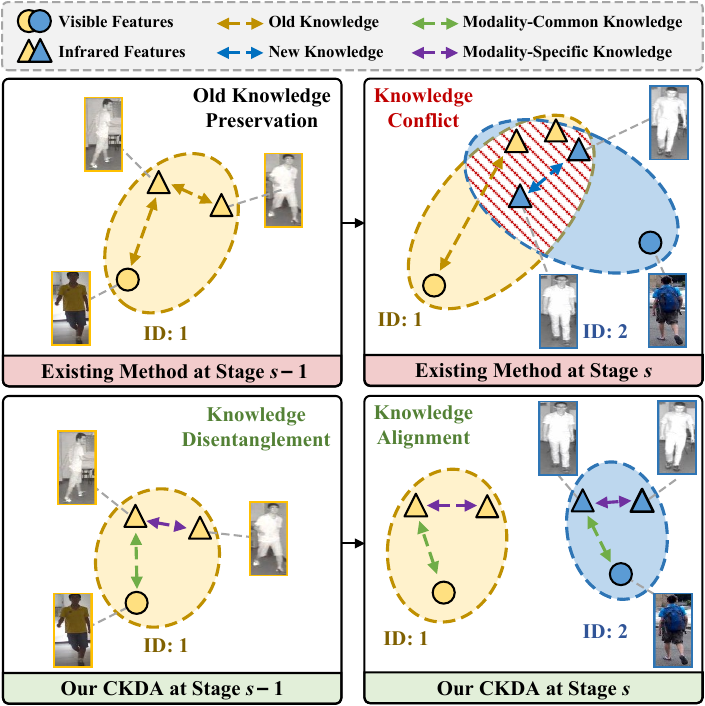}
\vspace{-5pt}
\caption{Comparison of different LReID methods when facing visible and infrared images. Existing methods suffer from the conflict between the new knowledge acquisition (\emph{e.g.}, acquiring radiation knowledge specific to the infrared modality) and the old knowledge preservation (\emph{e.g.}, preserving shape knowledge that coexists in both modalities, which conflicts with the former one). While our CKDA achieves knowledge balancing by explicitly aligning the disentangled modality-common and -specific knowledge.}
\vspace{-10pt}
\label{fig: motivation}
\end{figure}

\par Lifelong person Re-IDentification (LReID) aims to employ sequentially collected data to match the same person in all scenarios~\cite{pu2021lifelong, pu2023memorizing}. Its core challenge lies in mitigating catastrophic forgetting of old knowledge learned in known scenarios while continuously acquiring new knowledge from new scenarios~\cite{de2021continual, masana2022class, xu2024lstkc}. To this end, existing LReID methods have demonstrated their effectiveness in suppressing the forgetting of old knowledge through data replay~\cite{yu2023lifelong, ge2022lifelong} and knowledge distillation~\cite{sun2022patch, cui2024learning}. Despite some progress in visible modality, these methods typically suffer from a more practical and challenging scenario where both visible and infrared images are continuously collected to match a person across day and night, which is also called Visible-Infrared Lifelong person Re-IDentification (VI-LReID)~\cite{xing2024lifelong}.

\par As a preliminary pioneer, Visible-Infrared person Re-IDentification (VI-ReID) matches the same person captured by visible and infrared cameras by extracting discriminative information against the discrepancy of heterogeneous information within different modalities~\cite{ye2020dynamic, cui2024dma}. To this end, modality-common representation-based methods~\cite{wu2021discover, yu2023modality} extract features from different modalities and map them into a common feature space to align information of the same person, while modality-specific representation-based methods~\cite{ren2024implicit, yu2023modality} focus on extracting unique information from each modality to further enrich the diversity and distinctiveness of individual representations. Therefore, some recent works attempt to transfer the above methods to alleviate the knowledge forgetting of both modalities in VI-LReID~\cite{xing2024lifelong, luo2025lifelong}. Among them, some existing methods follow the data rehearsal paradigm and retain old knowledge by inference task matching~\cite{xing2024lifelong} and distilling cross-modal knowledge~\cite{luo2025lifelong} between old visible and infrared data. Inspired by prompting technology, the recent VI-LReID method~\cite{zhu2025lifelong} introduced a dynamic prompt selection mechanism to preserve instance-level old knowledge while learning the most relevant new knowledge for the current scenario from a modality-common prompt pool.

\par However, in addition to the privacy issue caused by the data rehearsal~\cite{ahmad2022event}, these methods typically ignore the conflict between modality-specific knowledge acquisition and modality-common knowledge preservation, making it difficult to balance the cross-modal knowledge. Specifically, as shown in Figure~\ref{fig: motivation}, when learning on new data, existing VI-LReID methods inevitably sacrifice the old knowledge that coexists between modalities (\emph{e.g.}, body shape knowledge) due to accumulating new knowledge that are specific to distinct modality (\emph{e.g.}, thermal radiation knowledge in infrared modality, which is useless or even harmful for matching thermal radiation information that does not exist in the visible modality). As a result, the conflicting modality-specific knowledge acquisition and modality-common knowledge preservation typically lead to joint forgetting due to their mutual interference, which further eliminates the discriminability of cross-modality knowledge in both new and old scenarios.

\par To tackle the above issue, this paper proposes a Cross-modality Knowledge Disentanglement and Alignment (CKDA) method for the VI-LReID task, which aims to balance the conflicting cross-modality knowledge acquisition and preservation. Specifically, to alleviate the mutual interference between cross-modal knowledge, we first proposed a Modality-Common Prompting (MCP) module and a Modality-Specific Prompting (MSP) module to disentangle both knowledge. The former aims to extract modality-common knowledge that coexists in both modalities, while the latter further purifies knowledge that only exists in visible or infrared modality in a complementary manner. In addition, a Cross-modality Knowledge Aligning (CKA) module is further designed to align the above knowledge with the constructed inter- and intra-modality feature space by employing identity-level prototypes of both modalities, avoiding the rehearsing of old data. In summary, the main contributions of this paper are as follows:

\begin{itemize}
    \item A Cross-modality Knowledge Disentanglement and Alignment (CKDA) method is proposed to tackle the Visible-Infrared Lifelong person Re-IDentification challenge, which alleviates catastrophic forgetting derived from the conflict between cross-modality knowledge acquisition and preservation.
    \item A Modality-Common Prompting (MCP) module and a Modality-Specific Prompting (MSP) module are designed to disentangle and purify modality-common and -specific knowledge, explicitly suppressing the mutual interference between both modalities.
    \item A Cross-modal Knowledge Alignment (CKA) module is designed to balance the disentangled knowledge with a pair of symmetric inter- and intra-modality feature spaces constructed by visible and infrared identity prototypes.
\end{itemize}

\begin{figure*}[ht]
\centering
\includegraphics[width=0.95\linewidth]{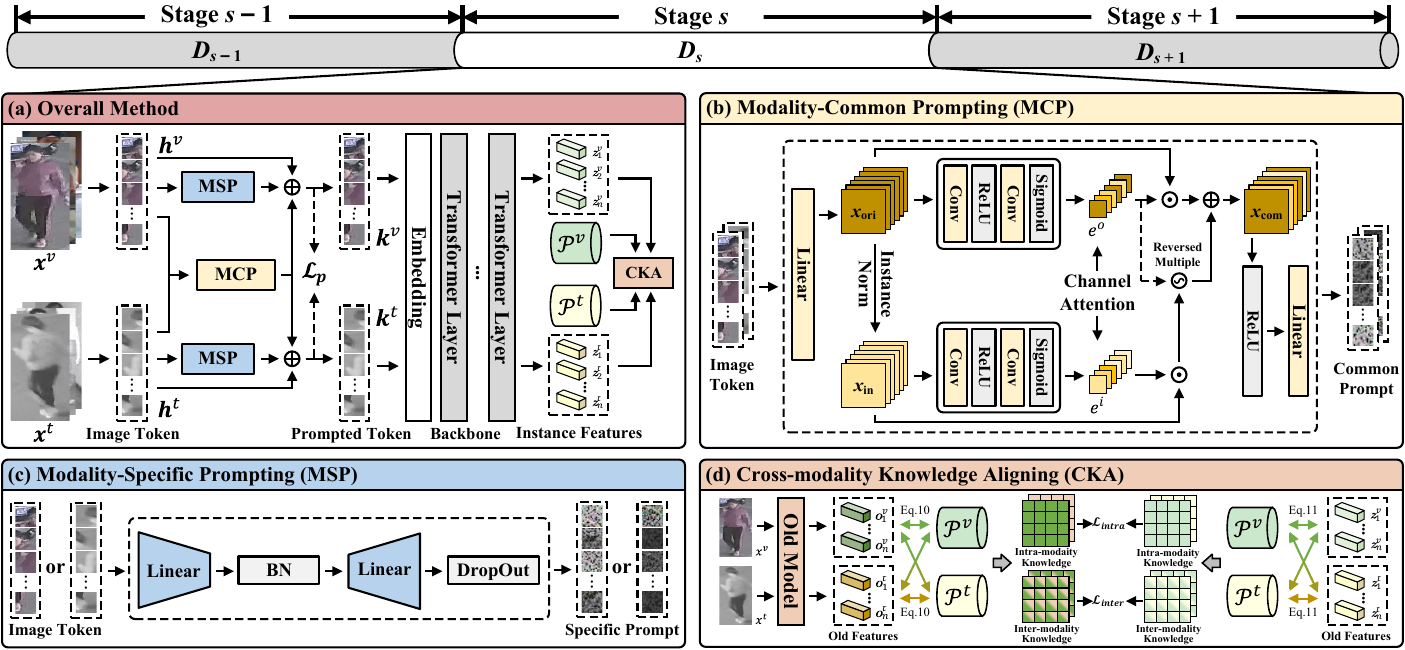}
\vspace{-5pt}
\caption{Overview of our proposed Cross-modality Knowledge Disentanglement and Alignment (CKDA) method. The input images are first fed into the Modality-Common Prompting (MCP) module and the Modality-Specific Prompting (MSP) module to generate the corresponding prompted image tokens. Then, the Cross-modality Knowledge Aligning (CKA) module exploits the cross-modality knowledge prototype to align the above two kinds of knowledge, respectively.}
\vspace{-10pt}
\label{fig: method}
\end{figure*}

\section{Related Work}

\subsection{Lifelong Person Re-identification}

\par Other than ReID in pre-collected data~\cite{cui2023dcr}, Lifelong person Re-IDentification (LReID) aims to employ sequentially collected data to match the same person in all scenarios~\cite{pu2021lifelong}. To this end, replay-based LReID methods rehearse old data when learning on new data to tackle the catastrophic forgetting challenge~\cite{wu2021generalising, ge2022lifelong, yu2023modality, chen2022unsupervised}, which raises concerns about the data privacy issue. In contrast, non-replay methods are dedicated to distilling from the old model to preserve old discriminative knowledge~\cite{xu2024lstkc, cui2024learning, xu2025dask}. In addition to the above works that rely on well-labelled data, some recent efforts have also explored lifelong person re-identification in more complex open scenarios, \emph{e.g.}, noisy~\cite{xu2024mitigate}, semi-supervised~\cite{xu2025self} and unsupervised scenarios~\cite{chen2022unsupervised}, further expanding its real significance.
\par However, existing LReID methods typically focus on knowledge balancing within visible data, ignoring the requirement of all-day pedestrian matching across day and night with images captured by visible and infrared cameras, respectively, which limits their applicability in practice.

\subsection{Visible-Infrared Person Re-identification}

\par Visible-Infrared Person Re-IDentification (VI-ReID) aims to match the same person across images captured by both infrared cameras and visible cameras~\cite{wu2017rgb}. To this end, modality-common representation-based methods~\cite{cui2024dma, liang2023cross, lu2023learning, ye2020dynamic, sun2023counterfactual, wu2021discover, park2021learning} extract and align common discriminative information that coexists in different modalities, while modality-specific representation-based methods~\cite{ren2024implicit, yu2023modality, li2022visible, hu2022adversarial, zhang2022fmcnet, jiang2022cross} focus on exploiting and leveraging information that only exists in a specific modality, achieving more comprehensive individual information representation.
\par Despite some progress, these methods typically suffer from severe performance degradation when new data is continuously learned in a streaming manner. Inspired by recent advancements in VI-ReID, some works~\cite{xing2024lifelong, luo2025lifelong} began to focus on a practical and challenging task, called Visible-Infrared Lifelong Person Re-IDentification (VI-LReID), which aims to avoid the catastrophic forgetting of cross-modality knowledge when learning from streaming data sequentially collected by visible and infrared cameras. Among them, TTQK~\cite{xing2024lifelong} learns a specific token for each domain, and matches the most appropriate inference token for each query sample to preserve old knowledge. In order to avoid the additional query token selection overhead, DMM~\cite{luo2025lifelong} alleviate the catastrophic forgetting problem by distilling the similarity between samples of different modalities. Inspired by recent advances in visual prompting~\cite{yao2025selective, zhou2025state, ai2025gaprompt, ai2025upp}, PP-IPG~\cite{zhu2025lifelong} further exploited fine-grained information of each modality at the instance level through a prompt pool. However, in addition to the privacy issue derived from the heavy reliance on old data, these methods typically suffer from the mutual interference of modality-specific knowledge acquisition and modality-common knowledge anti-forgetting, which limits their capacity to balance the cross-modal knowledge.
\par Different from these methods, we propose a Cross-modality Knowledge Disentanglement and Alignment (CKDA) method for VI-LReID, which avoids mutual interference between different knowledge by explicitly disentangling and adaptive balancing cross-modal knowledge through dual knowledge alignment.

\section{Method}

\subsection{Preliminary}

\par This paper focuses on Visible-Infrared Lifelong person Re-IDentification (VI-LReID) task. Formally, given a sequentially collected data stream $D=\{D_{1}, D_{2}, ..., \mathcal{D}_{S}\}$ with $S$ stages, where each dataset $\mathcal{D}_{s}$ consists of a visible image set $\mathcal{D}^{v}_{s}$ and an infrared image set $\mathcal{D}^{t}_{s}$. When sequentially learning on $\mathcal{D}_{s}$, the previous $s-1$ datasets are inapplicable due to the data privacy issue. For inference, the overall re-identification performance of each dataset between different modalities is employed for evaluation.

\subsection{Overview of CKDA}

\par The pipeline of our CKDA is shown in Figure~\ref{fig: method}, which consists of a Modality-Common Prompting (MCP) module, a Modality-Specific Prompting (MSP) module and a Cross-modality Knowledge Aligning (CKA) module. Then, we will elaborate on them in the following sections.

\noindent\textbf{Baseline.} Inspired by existing LReID methods~\cite{xu2024lstkc, xu2024distribution}, given a batch of input images with identities $\{\bm{x}_{i}, \bm{y}_{i}\}^B_{i=1}\in\mathcal{D}_{s}$ at stage $s$, we input each $\bm{x}_{i}$ into a backbone network $\phi_s(\cdot)$ with a batch normalization~\cite{ioffe2015batch} layer to extract the deep feature $\bm{z}_i$. Then, an identity classifier $\psi_s(\cdot)$ is exploit to predict the identity probability $\bm{y}'_{i}$ for each image $\bm{x}_{i}$.
\par To facilitate the acquisition of new knowledge, we introduce a classification loss $\mathcal{L}_{ce}$~\cite{luo2019bag} and a triplet loss $\mathcal{L}_{trip}$~\cite{hermans2017defense} to improve the discriminability of each person. Therefore, the overall optimization objective of our baseline is formulated as:
\begin{equation}
\label{eq:loss_base}
\begin{aligned}
\mathcal{L}_{base} = \mathcal{L}_{ce} + \mathcal{L}_{trip}.
\end{aligned}
\end{equation}
\par Besides, we adopt the Exponential Moving Average (EMA) strategy, which is widely used in existing LReID methods~\cite{xu2024distribution, xu2025dask} to initially avoid the forgetting of old knowledge caused by drastic parameter changes after each training stage as follows:
\begin{equation}
\label{eq:loss_ema}
\begin{aligned}
\phi_s = \lambda\cdot\phi_{s-1} + (1-\lambda)\cdot\phi_s^*,
\end{aligned}
\end{equation} where $\phi_s^*$ denotes the backbone trained after stage $s$, $\lambda$ denotes the balance factor between old and new knowledge.

\subsection{Modality-Common Prompting}

\par To explicitly disentangle the modality-common knowledge, we design a Modality-Common Prompting (MCP) module to capture the discriminative information that co-exists in visible and infrared modalities. Therefore, MCP adaptively incorporates the features after modality discrepancies elimination to purify modality-common information.

\par Specifically, given an input image $\bm{x}\in\mathbb{R}^{H\times W\times C}$, we first divide it into $M$ image tokens $\bm{h}=\{h_i\}_{i=1}^M\in \mathbb{R}^{ps\times ps\times pc}$ of the same size, where $H,W$ are the height and width of image $\bm{x}$, $ps$ is the size of each image token, $C$ and $pc$ is the channel number of $\bm{x}$ and $\bm{h}$. Each image token is then embedded into a $d$-dimension latent feature as follows:
\begin{equation}
\label{eq:MCP_linear_a}
\begin{aligned}
\bm{h}' = \mathcal{E}_d(\bm{h}),
\end{aligned}
\end{equation} where $\mathcal{E}_d(\cdot)$ denotes the embedding layer of the image token. Then, we rearrange $\bm{h}'$ to form a feature map $\bm{x}_{ori}\in\mathbb{R}^{H\times W\times C}$ for knowledge purification. 

\par To purify the modality-common knowledge, we exploit the Instance Normalization~\cite{ulyanov2016instance} to alleviate the style discrepancy between different modalities, while avoiding the derived loss of discriminative information. Specifically, We first obtain the normalized features $\bm{x}_{in}$ as follows:
\begin{equation}
\label{eq:MCP_in}
\begin{aligned}
\bm{x}_{in} = \frac{\bm{x}_{ori}-\mathrm{E}[\bm{x}_{ori}]}{\sqrt{\mathrm{Var}[\bm{x}_{ori}]+\epsilon}},
\end{aligned}
\end{equation} where the parameter $\epsilon$ is applied to avoid division by zero errors, $\mathrm{E}[\cdot]$ and $\mathrm{Var}[\cdot]$ denote the channel-wise mean and the variance, respectively. Then, we evaluate the modality-common knowledge distribution of $\bm{x}_{ori}$ and $\bm{x}_{in}$ by generating two channel-wise masks $e^o$ and $e^i$ as follows:
\begin{equation}
\label{eq:MCP_linear_att}
\begin{cases}
e^o=\sigma(W_2^o\cdot\delta(W_1^o\cdot \bm{x}_{ori})) \\
e^i=\sigma(W_2^i\cdot\delta(W_1^i\cdot \bm{x}_{in})) 
\end{cases},
\end{equation} where $\sigma(\cdot)$ and $\delta(\cdot)$ denote the ReLU and the sigmoid activation function, $W_1^o, W_1^i\in \mathbb{R}^{\frac{d}{r}\times d}$ and $W_2^o, W_2^i\in \mathbb{R}^{d \times\frac{d}{r}}$ denote two learnable parameters. Considering the balance between complexity and efficiency, the dimension reduction $r$ is set to 4. To further balance the discriminability and discrepancy of modality-common knowledge, MCP adaptively fuses $\bm{x}_{ori}$ and $\bm{x}_{in}$ as follows:
\begin{equation}
\label{eq:MCP_linear_fuse}
\bm{x}_{com}=e^o\cdot \bm{x}_{ori}+(1-e^o)(e^i\cdot \bm{x}_{in}),
\end{equation} where $e^i$ denotes the importance of modality-common knowledge after eliminating modality discrepancies, while $(1-e^o)$ supplements the discriminative information after suppressing the modality discrepancy in a dynamic manner according to the importance of the modality-common knowledge in the original feature $\bm{x}_{ori}$.

\par Finally, we align the common prompting $\bm{x}_com$ with the original feature map to generate the common prompt $\bm{k}_{com}$ by restoring the input dimension based on the feature token $\bm{h}''$, which is obtained by token division of $\bm{x}_{com}$ as follows:
\begin{equation}
\label{eq:MCP_linear_b}
\bm{k}_{com}=\mathcal{E}_{pc}(\delta(\bm{h}'')),
\end{equation} where $\mathcal{E}_{pc}(\cdot)$ denotes the embedding restoration layer. In this way, the modality-common knowledge can be dynamically purified in the unified common prompt, alleviating the mutual interference of both modalities.

\subsection{Modality-Specific Prompting}

\par Although the Modality-Common Prompting module disentangles and purifies the common knowledge between different modalities, it is still limited by the insufficient preservation of modality-specific knowledge, which further suppresses the balance of cross-modal knowledge. Therefore, we further propose a Modality-Specific Prompting (MSP) module to further disentangle the modality-specific discriminative information that only exists in a distinct modality.

\par To this end, a pair of light-weight modality-specific prompt generation networks is designed for visible and infrared modalities, respectively. Specifically, given an image token $\bm{h}^m$ of a specific modality, where $m\in\{v, t\}$ represents the visible or the infrared modality. The modality-specific prompt $\bm{k}_{spe}$ can be calculated as follows:
\begin{equation}
\label{eq:MSP}
\bm{k}^m_{spe}=\xi(W_2^m\cdot\theta(W_1^m\cdot \bm{h}^m)),
\end{equation} where $\xi(\cdot)$ denotes the dropout layer, $W_1^m\in \mathbb{R}^{\frac{d}{r} \times d}$, $W_2^m\in \mathbb{R}^{d \times\frac{d}{r}}$ denote the learnable parameters of the linear layer, $\theta(\cdot)$ denotes the Batch Normalization layer. Consequently, the modality-specific prompting is encouraged to further stimulate the difference between modalities with batch-wise regularization, thereby facilitating the model to extract modality-specific information in distinct modalities under explicit modality discrepancies.

\noindent \textbf{Optimization of Modality-Com and MSP module.} To alleviate the catastrophic forgetting of cross-modal knowledge due to the continual knowledge change across visible and infrared data of different scenarios, we align the generated prompting of the current stage (Stage $s$) with that of the previous stage (Stage $s-1$). Let $\bm{k}^{m,(s)}_p=\bm{k}^{m,(s)}_{spe}+\bm{k}^{(s)}_{com}$ denote the image prompt at Stage $s$, the optimization can be achieved by minimizing the prompting loss $\mathcal{L}_p$, which can be calculated as follows:
\begin{equation}
\label{eq:L_p}
\mathcal{L}_p=|\bm{k}^{m,(s)}_p-\bm{k}^{m,(s-1)}_p|.
\end{equation}

\par Finally, we can get the prompted image $\bm{k}^m$ by merging the original image $\bm{x}^m$ and $\bm{k}^m_p$ at the current stage, which is then fed into the backbone network for feature extraction.

\begin{table*}[htbp]
\renewcommand{\arraystretch}{1}
  \centering
  \small
  \setlength{\tabcolsep}{1.1mm}{
    \begin{tabular}{c|l|c|c|cc|cc|cc|cc||>{\columncolor{gray!20}}c>{\columncolor{gray!20}}c}
	\hline
	\multicolumn{2}{c|}{\multirow{2}{*}{Method}} & \multirow{2}{*}{Publication} & \multirow{2}{*}{Backbone} & \multicolumn{2}{c|}{RegDB} & \multicolumn{2}{c|}{SYSU-MM01} & \multicolumn{2}{c|}{LLCM} & \multicolumn{2}{c||}{VCM} & \multicolumn{2}{>{\columncolor{gray!20}}c}{Average}\\
    \hhline{~~~~----------}
	\multicolumn{2}{c|}{}   & & & mAP   & R1   & mAP   & R1   & mAP   & R1   & mAP   & R1   & mAP   & R1\\
	\hline
	\multicolumn{2}{c|}{JointTrain} & - & ViT-B/16	& 74.8	& 79.4	& 55.8	& 59.4	& 60.2	& 56.3	& 19.1	& 33.4	& 52.5	& 57.1 \\
	\multicolumn{2}{c|}{SFT}	& - & ViT-B/16	& 6.8	& 5.7	& 21.0	& 21.5	& 26.4	& 21.3	& 17.1	& 29.0	& 17.8	& 19.4\\
    \hline
    
	\multirow{6}{*}{\rotatebox{90}{Replay}}	& LwF~\cite{li2017learning}	& \textcolor{gray}{\emph{T-PAMI}}	& ViT-B/16	& 26.9	& 26.1	& 30.9	& 30.3	& 39.1	& 34.5	& 18.2	& 30.0	& 28.8	& 30.2\\
	& iCaRL~\cite{rebuffi2017icarl}	& \textcolor{gray}{\emph{CVPR}}	& ViT-B/16	& 42.4	& 41.3	& 32.4	& 32.4	& 41.9	& 36.8	& 18.5	& 30.7	& 33.8	& 35.3\\
	& BiC~\cite{wu2019large}	& \textcolor{gray}{\emph{CVPR}}	& ViT-B/16	& 40.9	& 39.3	& 22.8	& 21.9	& 43.3	& 37.5	& 16.4	& 27.4	& 30.9	& 31.5\\
	& WA~\cite{zhao2020maintaining}	& \textcolor{gray}{\emph{CVPR}}	& ViT-B/16	& 44.9	& 43.9	& 34.6	& 34.4	& 44.1	& 38.7	& 18.7	& 29.3	& 35.6	& 36.6\\
	& PTKP~\cite{ge2022lifelong}	& \textcolor{gray}{\emph{AAAI}}	& ViT-B/16	& 50.3	& 47.9	& 26.6	& 25.8	& 43.2	& 36.5	& 17.9	& 27.4	& 34.5	& 34.4\\
	& TTQK~\cite{xing2024lifelong}	& \textcolor{gray}{\emph{ICMR}}	& ViT-B/16	& 60.3	& 58.3	& 33.6	& 30.9	& 45.8	& 39.6	& 19.4	& 32.2	& 39.8	& 40.2\\
	& PP-IPG~\cite{luo2025lifelong}	& \textcolor{gray}{\emph{ICMR}}	& ViT-B/16	& 59.0	& 58.8	& 32.7	& 35.2	& 44.4	& 50.8	& 35.3	& 23.3	& 42.8	& 42.0\\
    \hline
	\multirow{12}{*}{\rotatebox{90}{Non-Replay}}	& LwF~\cite{li2017learning}	& \textcolor{gray}{\emph{T-PAMI}}	& ResNet	& 23.4	& 23.0	& 9.5	& 9.7	& 17.8	& 13.6	& 6.8	& 11.6	& 14.4	& 14.6\\
	& AKA~\cite{pu2021lifelong}	& \textcolor{gray}{\emph{CVPR}}	& ResNet	& 26.2	& 23.9	& 11.0	& 10.0	& 17.2	& 12.9	& 7.4	& 13.0	& 15.5	& 15.1\\
	& PatchKD~\cite{sun2022patch}	& \textcolor{gray}{\emph{MM}}	& ResNet	& 52.2	& 55.2	& 9.6	& 7.9	& 18.5	& 13.7	& 3.9	& 5.9	& 21.1	& 20.8\\
	& LSTKC~\cite{xu2024lstkc}	& \textcolor{gray}{\emph{AAAI}}	& ResNet	& 15.9	& 13.5	& 15.1	& 13.4	& 22.1	& 17.4	& 16.0	& 26.8	& 17.3	& 17.8\\
	& DKP~\cite{xu2024distribution}	& \textcolor{gray}{\emph{CVPR}}	& ResNet	& 34.1	& 33.5	& 26.4	& 27.1	& 32.8	& 27.9	& 15.3	& 26.4	& 27.2	& 28.7\\
	& DASK~\cite{xu2025dask}	& \textcolor{gray}{\emph{AAAI}}	& ResNet	& 17.3	& 15.4	& 23.6	& 22.0	& 29.1	& 24.1	& 18.1	& 25.9	& 22.0	& 21.8\\
	& LSTKC++~\cite{xu2025long}	& \textcolor{gray}{\emph{T-PAMI}}	& ResNet	& 36.1	& 38.1	& 35.7	& 37.0	& 29.3	& 24.9	& 16.3	& 26.6	& 29.4	& 31.6\\
    \hhline{~-------------}
    & LSTKC~\cite{xu2024lstkc}	& \textcolor{gray}{\emph{AAAI}}	& ViT-B/16	& 35.0	& 35.4	& 32.7	& 34.3	& 39.6	& 35.1	& 15.1	& 25.3	& \textbf{\textcolor{blue}{30.6}}	& \textbf{\textcolor{blue}{32.5}}\\
	& DKP~\cite{xu2024distribution}	& \textcolor{gray}{\emph{CVPR}}	& ViT-B/16	& 30.3	& 32.7	& 30.9	& 32.8	& 37.3	& 32.2	& 15.2	& 26.3	& 28.4	& 31.0\\
    & TTQK~\cite{xing2024lifelong}  & \textcolor{gray}{\emph{CVPR}} & ViT-B/16  & 8.1   & 6.5   & 20.8  & 18.6  & 19.2  & 14.9  & 20.8  & 32.6  & 17.2  & 18.2\\
	& DASK~\cite{xu2025dask}	& \textcolor{gray}{\emph{AAAI}}	& ViT-B/16	& 21.6	& 25.6	& 28.1	& 32.3	& 39.6	& 36.2	& 14.1	& 27.4	& 25.9	& 30.4\\
	& LSTKC++~\cite{xu2025long}	& \textcolor{gray}{\emph{T-PAMI}}	& ViT-B/16	& 19.2	& 19.0	& 27.1	& 30.3	& 32.9	& 29.2	& 16.1	& 29.2	& 23.8	& 26.9\\
    \hhline{~-------------}
	\cellcolor{white}	& \textbf{Ours}	& This Paper	& ViT-B/16	& 57.0	& 60.4	& 34.7	& 37.3	& 37.8	& 31.5	& 15.6	& 28.2	& \textbf{\textcolor{red}{36.3}}	& \textbf{\textcolor{red}{39.4}}\\
    \hline
    \end{tabular}
    }
    \vspace{-8pt}
    \caption{Performance Comparison on training order: RegDB$\rightarrow$SYSU-MM01$\rightarrow$LLCM$\rightarrow$VCM.
    }
    \vspace{-6pt}
  \label{tab: setting1}%
\end{table*}%

\subsection{Cross-modality Knowledge Aligning}

\par Although the above modules eliminate the mutual interference between different knowledge by disentanglement, it is still challenging to simultaneously alleviate two conflicting knowledge in a unified feature space. Therefore, we further propose a Cross-modality Knowledge Aligning (CKA) module, which aims to align modality-specific knowledge in an intra-modality feature space while aligning modality common knowledge in a dual inter-modality feature space.

\par To this end, we employ the feature prototype as the basis of the old feature space to measure and align cross-modal knowledge. Specifically, given a pair of visible prototypes $\mathcal{P}^v_{s-1}=\{p^v_i\}_{i=1}^{N_{s-1}}$ and infrared prototypes $\mathcal{P}^t_{s-1}=\{p^t_i\}_{i=1}^{N_{s-1}}$, where $N_{s-1}$ denote the identity number of the $(s-1)^{th}$ stage, $p^v_i$ and $p^t_i$ denote the visible and infrared feature center of old data extracted by $\phi_{s-1}(\cdot)$, we use the current visible instance features $\{o_i^v\}^{N_s}_{i=1}$ and infrared instance features $\{o_i^t\}^{N_s}_{i=1}$ extracted by $\phi_{s-1}(\cdot)$ to represent the old inter-modality feature space $\bm{O}$ as follows (taking visible-to-infrared as an example):
\begin{equation}
\label{eq:CKA_old}
\begin{aligned}
\bm{O}_{vt}(i,j) = \displaystyle{\frac{\text{exp}(cos(\bm{o}^{v}_i,\bm{p}^{t}_j)/\tau)}{\sum_{k=1}^{N_{s-1}}\text{exp}(cos(\bm{o}^{t}_i,\bm{p}^{v}_k)/\tau)}},
\end{aligned}
\end{equation} where $cos(\cdot)$ denotes the cosine similarity, $\tau$ is a temperature parameter set as 0.1. Thus, the intra-modality feature space can be represented by $\bm{O}_{vv}$ and $\bm{O}_{tt}$, which is employed to measure the modality-common knowledge distribution under the old intra-modality feature space. Similarly, we can further exploit the visible features $\{z_i^v\}^{N_s}_{i=1}$ and infrared features $\{z_i^t\}^{N_s}_{i=1}$ extracted by the current model to represent the new distribution $\bm{Z}$ after the new knowledge acquisition: 
\begin{equation}
\label{eq:CKA_new}
\begin{aligned}
\bm{Z}_{vt}(i,j) = \displaystyle{\frac{\text{exp}(cos(\bm{z}^{v}_i,\bm{p}^{t}_j)/\tau)}{\sum_{k=1}^{N_{s-1}}\text{exp}(cos(\bm{z}^{t}_i,\bm{p}^{v}_k)/\tau)}}.
\end{aligned}
\end{equation}

\par Then, to alleviate the catastrophic forgetting of inter- and intra-modality knowledge, we align the inter- and intra-affinity of each feature with the proposed knowledge balancing loss $\mathcal{L}_{inter}$ and $\mathcal{L}_{intra}$ as follows:
\begin{equation}
\label{eq:CKA_loss_1}
\begin{aligned}
\mathcal{L}_{inter}=\mathcal{L}_{KL}(Y^o_{v,t}||Y^z_{v,t})+\mathcal{L}_{KL}(Y^o_{t,v}||Y^z_{t,v}),
\end{aligned}
\end{equation}
\begin{equation}
\label{eq:CKA_loss_2}
\begin{aligned}
\mathcal{L}_{intra}=\mathcal{L}_{KL}(Y^o_{v,v}||Y^z_{v,v})+\mathcal{L}_{KL}(Y^o_{t,t}||Y^z_{t,t}),
\end{aligned}
\end{equation}
\begin{equation}
\label{eq:CKA_loss_3}
\begin{aligned}
Y^o_{v,t}=\rho(\bm{O}_{vt}\bm{O}_{vt}^\top/\tau),\ \ Y^z_{v,t}=\rho(\bm{Z}_{vt}\bm{Z}_{vt}^\top/\tau),
\end{aligned}
\end{equation} where $\mathcal{L}_{KL}$ is the Kullback-Leibler (KL) divergence, and $\rho(\cdot)$ is the softmax function. In summary, the CKA module employs the old prototype as the old feature space to alleviate the cross-modal knowledge shift derived by the acquisition of new knowledge via dual knowledge alignment, thereby balancing the discriminative information between and within visible and infrared modalities.

\subsection{Overall Optimization}
\par The overall optimization objective for our CDKA can be formalized as follows:
\begin{equation}
\label{eq:loss}
\begin{aligned}
\mathcal{L} = \mathcal{L}_{base}+\alpha L_p+\beta(\mu{L}_{inter}+(1-\mu)\mathcal{L}_{intra}),
\end{aligned}
\end{equation} where $\alpha$ and $\beta$ denote two hyperparameters when training the model, $\mu$ denotes a scaling factor to balance the modality-specific and -common knowledge preservation.

\section{Experiments}

\subsection{Datasets and Evaluation Metrics}
\noindent\textbf{Datasets.} To verify the effectiveness of our CKDA, we follow the most popular VI-LReID benchmark in~\cite{xing2024lifelong}, which consists of 4 widely-used visible-infrared person re-identification datasets, including RegDB~\cite{nguyen2017person}, SYSU-MM01~\cite{wu2017rgb}, LLCM~\cite{zhang2023diverse}, and HITSZ-VCM (VCM)~\cite{lin2022learning}. To perform fairly comparison in the real lifelong scenario, we also follow~\cite{xing2024lifelong} and sequentially conduct the above datasets as the training order: RegDB$\rightarrow$SYSU-MM01$\rightarrow$LLCM$\rightarrow$VCM. For the video dataset VCM, we follow the frame-level sampling strategy in~\cite{xing2024lifelong} to generate visible and infrared images.

\noindent\textbf{Evaluation Metrics.} To compare our CKDA with existing methods, three popular metrics, including Rank-1 accuracy (R1)~\cite{moon2001computational}, mean Average Precision (mAP)~\cite{market1501} with their Average Forgetting (AF)~\cite{chaudhry2018riemannian}, are employed to verify the overall performance on all datasets. The former two metrics evaluate the ReID performance, while the last metric describes their forgetting rate via the drop value.

\begin{table}[tbp]
\renewcommand{\arraystretch}{1.0}
\small
    \centering
    \setlength{\tabcolsep}{1.1mm}{
    \begin{tabular}{c|c|c|c|c|c|c}
    \hline
        Method & TTQK & DASK & DKP & LSTKC & LSTKC++ & \cellcolor{gray!20}\textbf{Ours}\\
    \hline
        AF(mAP) & 36.5 & 15.1 & 12.1 & 8.7 & 8.5 & \cellcolor{gray!20}\textbf{5.1}\\
    \hline
        AF(R1) & 35.5 & 12.1 & 12.2 & 7.7 & 6.9 & \cellcolor{gray!20}\textbf{4.9}\\
    \hline
    \end{tabular}
    }
    \vspace{-8pt}
    \caption{AF comparison to non-replay methods.}
    \vspace{-10pt}
    \label{tab: af}
\end{table}

\subsection{Implementation Details}
\par We implement our CKDA on NVIDIA A40 GPU for both training and inference. During the training phase, ViT-B/16~\cite{dosovitskiy2020image} pre-trained on ImageNet~\cite{russakovsky2015imagenet} is employed as the initial backbone network. For each dataset, we randomly sample 500 identities and train the model for 50 epochs, where inadequate‌ datasets sample all identities. The batch size is set to 64, where each identity has 4 visible images and 4 infrared images. The learning rate is set to $3\times10^{-4}$ and a cosine decay schedule~\cite{chen2020simple, loshchilov2016sgdr} to optimize the whole model. During the inference phase, following~\cite{xing2024lifelong}, we perform visible-to-infrared re-identification on the RegDB datasets, and perform infrared-to-visible re-identification on the others.

\begin{table}[tbp]
\renewcommand{\arraystretch}{1.0}
\small
    \centering
    \setlength{\tabcolsep}{3.6mm}{
    \begin{tabular}{c|c|c|c|c|c}
    \hline
        Base & MCP & MSP & CKA & mAP & R1\\
    \hline
        \checkmark & - & - & - & 31.8 & 33.9\\
        \checkmark & \checkmark & - & - & 33.4 & 35.2\\
        \checkmark & \checkmark & \checkmark & - & 34.6 & 37.4\\
        \checkmark & - & - & \checkmark & 34.9 & 37.9 \\
        \hline
        \rowcolor{gray!20} \checkmark & \checkmark & \checkmark & \checkmark & \textbf{36.3} & \textbf{39.4}\\
    \hline
    \end{tabular}}
    \vspace{-8pt}
    \caption{Ablation study of each module in CKDA.}
    \vspace{-12pt}
    \label{tab: ablation}
\end{table}

\subsection{Comparison with State-of-the-arts Methods}

\par We start by compare our proposed CKDA to multiple SOTA methods on both LReID and VI-LReID tasks, including LwF~\cite{li2017learning}, iCaRL~\cite{rebuffi2017icarl}, BiC~\cite{wu2019large}, WA~\cite{zhao2020maintaining}, PTKP~\cite{ge2022lifelong}, TTQK~\cite{xing2024lifelong}, PP-IPG~\cite{luo2025lifelong}, AKA~\cite{pu2021lifelong}, PatchKD~\cite{sun2022patch}, LSTKC~\cite{xu2024lstkc}, DKP~\cite{xu2024distribution}, DASK~\cite{xu2025dask}, and LSTKC++~\cite{xu2025long}. Note that the best results are marked in \textbf{red}, while the sub-optimal results are marked in \textbf{blue}.

\par As shown in Table~\ref{tab: setting1}, our CKDA achieves \textbf{36.3\%}/\textbf{39.4\%} on the average mAP/R1 accuracy, which outperforms the existing non-replay SOTA method LSTKC~\cite{xu2024lstkc} by \textbf{5.7\%}/\textbf{6.9\%}. This is because our CKDA disentangles modality-specific knowledge from modality-common knowledge and thus eliminates their mutual interference. Besides, our method achieves effective dual preservation of both disentangled knowledge by dynamically aligning them in the modality-inter and -intra feature space in a balanced manner, thereby significantly outperforming the state-of-the-art methods.

\begin{figure}[ht]
\centering
\subfloat[The weight for $\alpha$]{\includegraphics[width=0.22\textwidth]{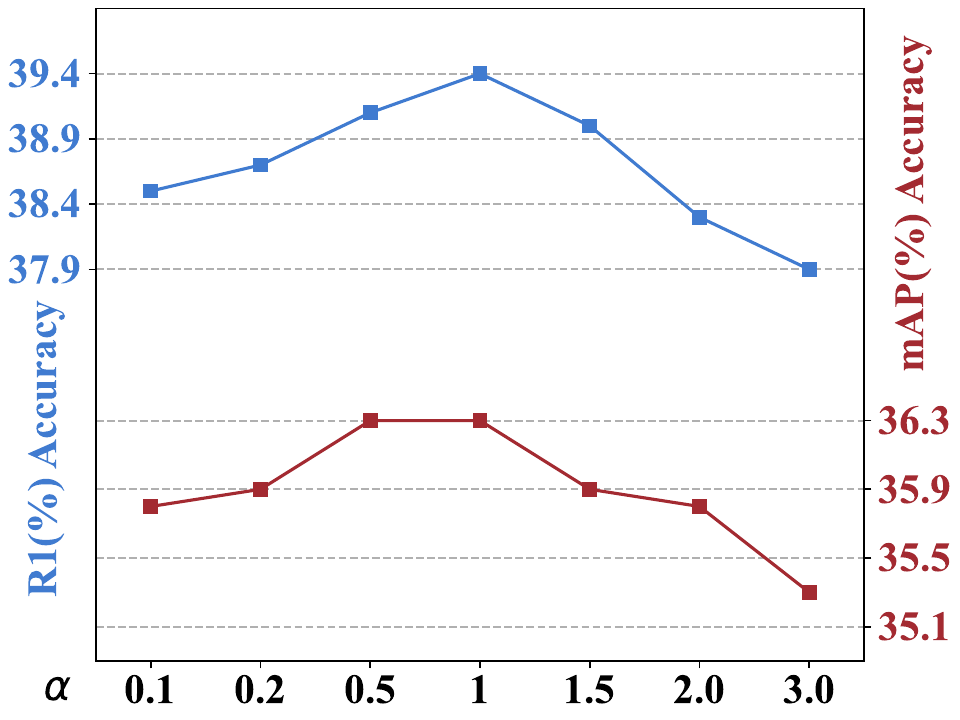}}
\hspace{1mm}
\subfloat[The weight for $\beta$]{\includegraphics[width=0.22\textwidth]{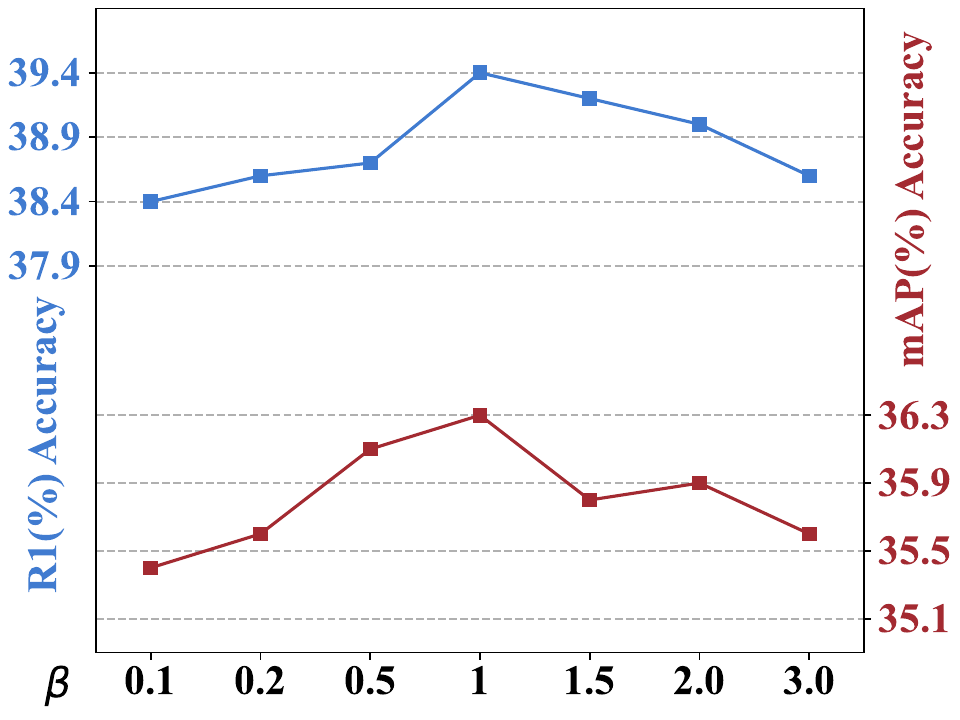}}
\hfill
\subfloat[The weight for $\mu$]{\includegraphics[width=0.22\textwidth]{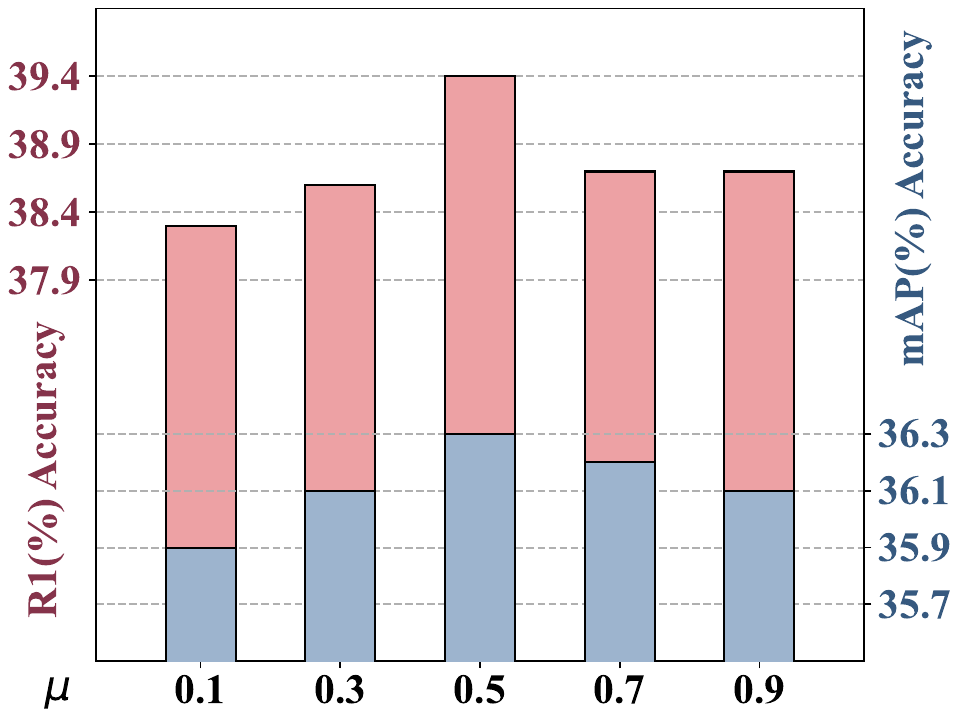}}
\hspace{1mm}
\subfloat[The weight for $d$]{\includegraphics[width=0.22\textwidth]{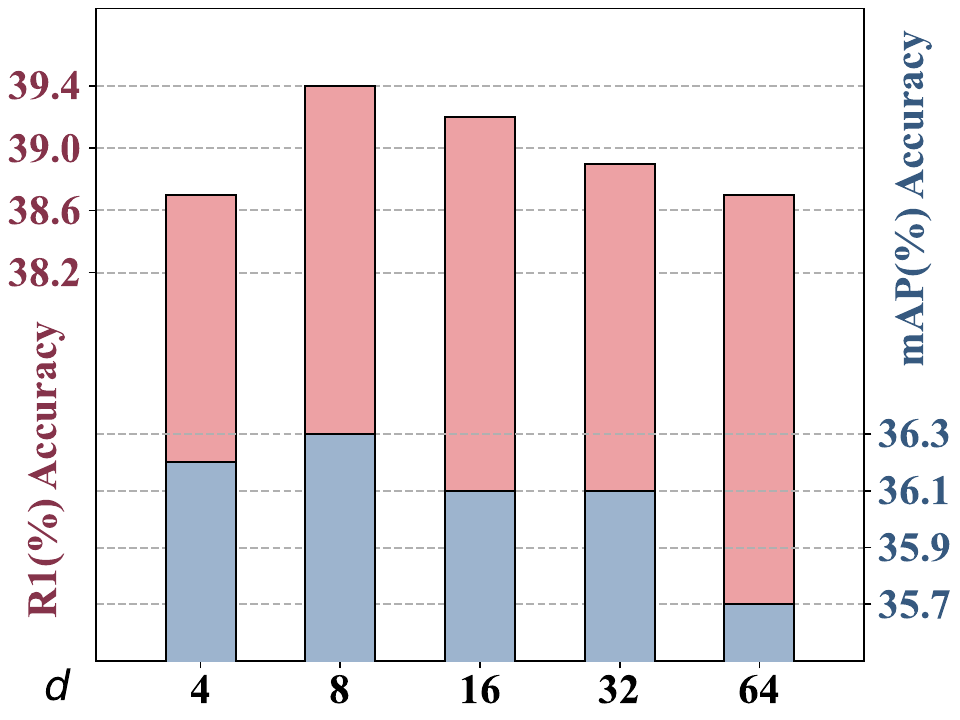}}
\vspace{-8pt}
\caption{The influence of hyperparameters in our CKDA.}
\vspace{-6pt}
\label{fig: params}
\end{figure}

\par Besides, Table~\ref{tab: af} reports the forgetting degree of our method, which achieves the best average anti-forgetting performance for \textbf{5.1\%}/\textbf{4.9\%} on the average AF(mAP/R1) performance. It implies that our proposed dual modality prompting splits the originally entangled cross-modal knowledge in an explicit way, and further preserves the discriminability of old knowledge through a pair of symmetric inter- and intra-modality distillations. Therefore, our method effectively alleviates the catastrophic forgetting derived from conflicting cross-modal knowledge.

\begin{figure}[ht]
\centering
\includegraphics[width=0.99\linewidth]{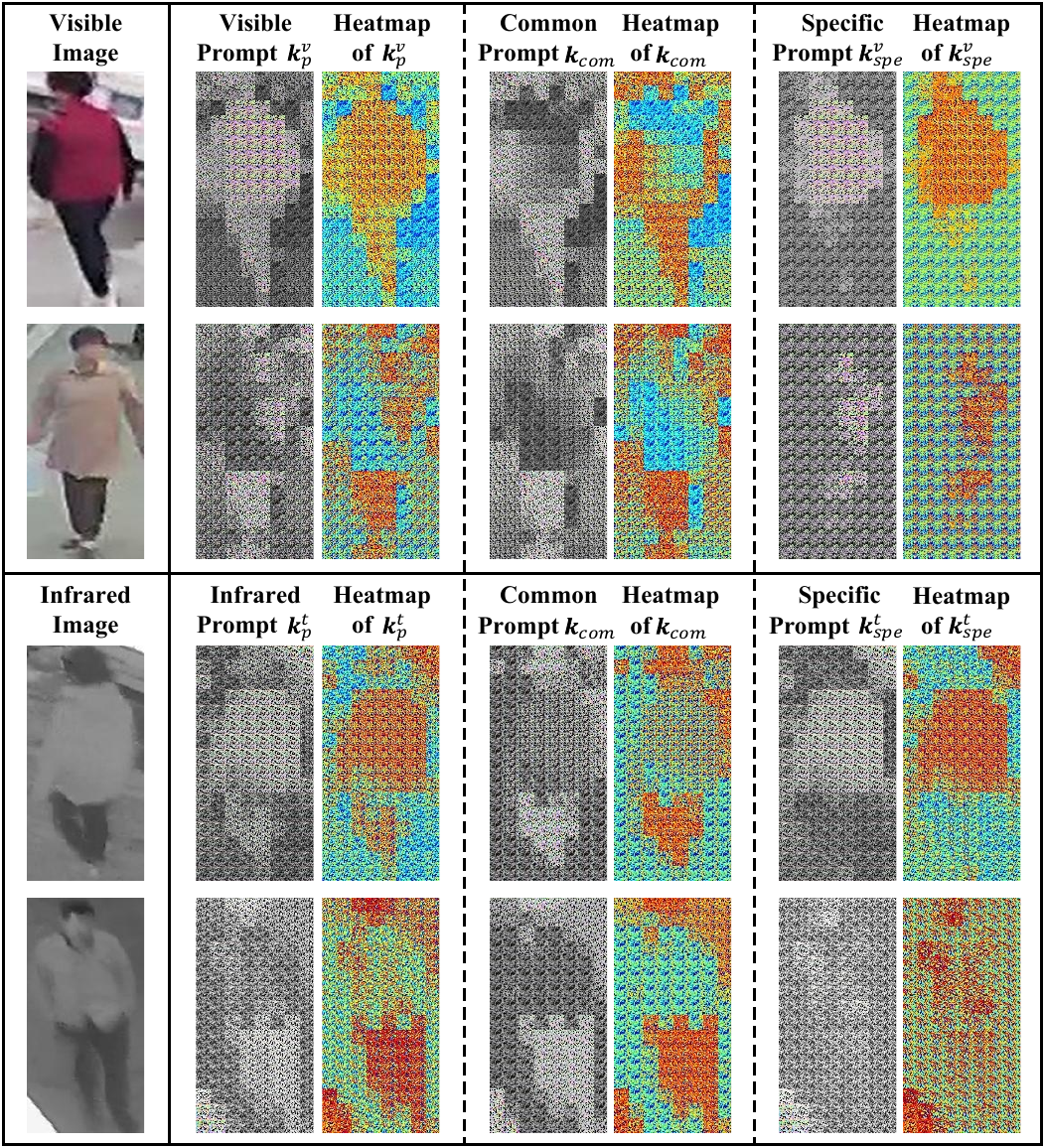}
\vspace{-6pt}
\caption{Visualization of the generated common prompt $\bm{k}_{com}$, specific prompt $\bm{k}_{spe}$, visible image prompt $\bm{k}^{v}$), and infrared image prompt $\bm{k}^{v}$.}
\vspace{-10pt}
\label{fig: heatmap}
\end{figure}

\begin{figure}[ht]
\centering
\includegraphics[width=0.99\linewidth]{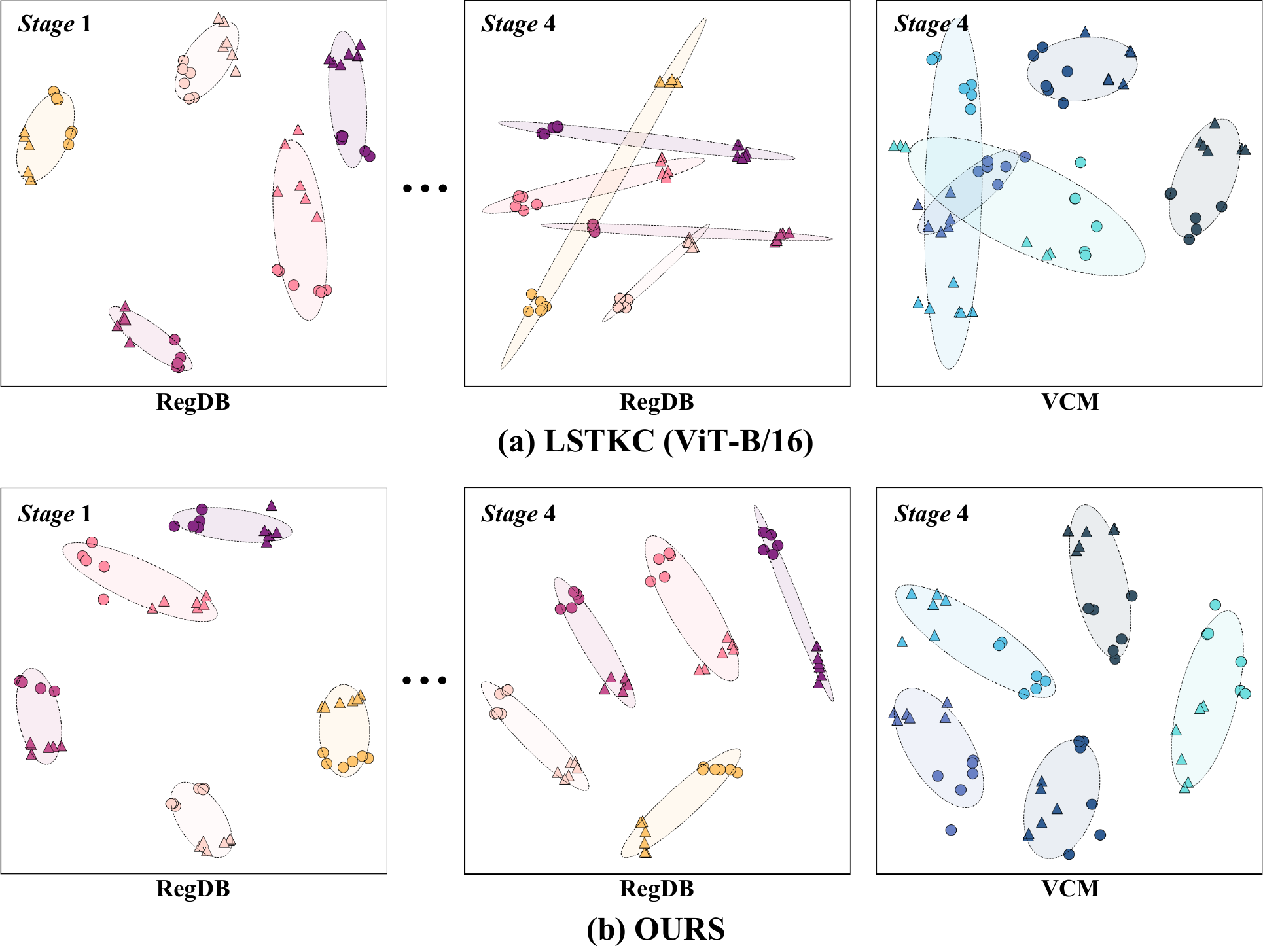}
\vspace{-6pt}
\caption{The t-SNE visualization of different training stages, where different colours represent different identities and $\bigcirc, \triangle$ represent features of infrared and visible modalities, respectively.}
\vspace{-10pt}
\label{fig: tsne}
\end{figure}

\subsection{Ablation Studies}

In this section, we conduct ablation studies to verify the effectiveness of each CKDA component.

\noindent\textbf{Effectiveness of the MCP module.} Table~\ref{tab: ablation} reports that MCP module brings \textbf{1.6\%}/\textbf{1.3\%} on average mAP and R1 accuracy compared to base method. It indicates that our approach effectively purifies the discriminative information that coexists in visible and infrared images by suppressing the discrepancy derived from distinct modality styles, thereby improving the adaptability of modality-common knowledge to both modalities.

\noindent\textbf{Effectiveness of the MSP module.} Different from the MCP module, the MSP module focuses on extracting discriminative information that only exists in distinct modalities. As shown in Table~\ref{tab: ablation}, MSP further improves mAP/R1 by \textbf{1.2\%}/\textbf{2.2\%}, which supports its further separation and complements the modality-specific discriminative information missing from the MCP module. Nevertheless, MSP does not sacrifice the modality-common knowledge introduced by MCP through encouraging distinct discriminative knowledge after amplifying the discrepancies between modalities, thus achieving disentanglement and purification of modality-common and -specific knowledge.

\noindent\textbf{Effectiveness of the CKA module.} The introduction of CKA brings \textbf{3.1\%}/\textbf{4.0\%} on mAP/R1 performance improvement, which is further enhanced to \textbf{4.5\%}/\textbf{5.5\%} when utilizing MCP and MSP modules together. It implies that aligning the disentangled cross-modal knowledge in two independent inter- and intra-modality feature spaces can effectively preserve the discriminability of multiple knowledge in a controllable manner. Thus, the originally conflicting modality-common and -specific knowledge can coexist in the lifelong process, alleviating the catastrophic forgetting derived from their mutual interference.

\noindent\textbf{Hyperparameter Analysis.} Our proposed CKDA introduces 4 hyper-parameters: $\alpha$, $\beta$, $\mu$ in Eq.~\ref{eq:loss}, and $d$ in Eq.~\ref{eq:MCP_linear_a}. Therefore, we evaluate their effects as shown in Figure~\ref{fig: params}. For $\alpha$ and $\beta$, our CKDA achieves the highest performance when both are set to 1. This is because an excessively large $\alpha$ prevents the model from separating the modality-common and -specific knowledge of the current data, while an excessively large $\beta$ over-aligns the new and old knowledge, suppressing the discriminability of the new knowledge. Similarly, $\mu$ controls the alignment degree of modality-common and -specific knowledge, where the best performance is achieved at $0.5$. This suggests a balance of equal importance between cross-modal knowledge, which further verifies that our CKDA can effectively separate the above knowledge in a balanced manner. Finally, the performance keeps improving until $d=8$ and gradually decreases for larger values, which indicates the trade-off between the redundancy and difficulty of cross-modal knowledge, while our method balances them and selects the most appropriate number of channels. Therefore, according to the above analysis, the hyper-parameters $\alpha$, $\beta$, $\mu$, and $d$ are set as 1, 1, 0.5 and 8, respectively.

\subsection{Visualization}
\par To intuitively demonstrate the effect of our method, we visualize the modality-common/-specific prompts with corresponding heatmaps as shown in Figure~\ref{fig: heatmap}. It reveals that the common prompt tends to focus on the overall contour and body shape information of the person, which coexist in both modalities. In contrast, the specific prompt turns to focus on modality-specific knowledge that is only available in distinct modalities, \emph{e.g.}, clothing color in visible images or heat-sensitive information in infrared images. As a result, the combined visible/infrared prompt can provide various perceptions in a complementary manner.
\par Additionally, Figure~\ref{fig: tsne} visualizes the feature distribution extracted by our CKDA compared to LSTKC~\cite{xu2024lstkc} at different stages, including the old dataset RegDB and the new dataset VCM. Limited by mutual interference, LSTKC fail to alleviate the conflict between modality-common and -specific knowledge, thus leading to severe dual knowledge forgetting. While our CKDA not only promotes the purification of cross-modality knowledge by explicit disentangling, but also facilitates old knowledge retention by aligning it in a balanced way. Therefore, our CKDA achieves leadership in both new knowledge acquisition and old knowledge retention.

\section{Conclusion}
\par In this paper, we focus on a practical and challenging problem called Visible-Infrared Lifelong person Re-IDentification (VI-LReID), which requires retrieving the same person based on sequentially collected cloth-consistent and cloth-changing data. To tackle the above issue, we propose a Modality-Common Prompting (MCP) module to balance cloth-relevant and cloth-irrelevant knowledge. Specifically, a Modality-Specific Prompting (MSP) module is designed to dynamically exploit differentiated new knowledge while eliminating the derived domain shift of old knowledge. Besides, a Cross-modality Knowledge Aligning (CKA) module is designed to alleviate the knowledge conflict between differentiated knowledge by aligning their distribution at different levels. Extensive experimental results illustrate that our CKDA outperforms existing methods in the challenging VI-LReID task.

\section*{Acknowledgements}

This work was supported by the grants from the National Natural Science Foundation of China (62525201, 62132001, 62432001) and Beijing Natural Science Foundation (L247006, L257005).

\small
\bibliography{aaai2026}

\end{document}